\documentclass[journal]{IEEEtran}
\ifCLASSINFOpdf
\else
\fi
\usepackage{graphicx}
\usepackage{subfigure}
\usepackage{tabularx}
\usepackage{amsthm}
\usepackage{multirow}
\usepackage{verbatim}
\usepackage{makecell}
\newcolumntype{Y}{>{\centering\arraybackslash}X}
\usepackage{array}
\usepackage[colorlinks,linkcolor=blue]{hyperref}
\begin{document}
%
\title{Attentive WaveBlock: Complementarity-enhanced Mutual Networks for Unsupervised Domain Adaptation in Person Re-identification and Beyond}
%
%
%

\author{Wenhao~Wang, Fang~Zhao, 
        Shengcai~Liao,~\IEEEmembership{Senior Member,~IEEE,}
        and~Ling~Shao,~\IEEEmembership{Fellow,~IEEE}
\thanks{W. Wang is with the School of Mathematical Sciences, Beihang University, Beijing, China. He is also with ReLER, University of Technology Sydney, Sydney, Australia. He finished his part of work during his internship in IIAI.}
\thanks{F. Zhao is with the Tencent AI Lab, Shenzhen, China.}
\thanks{S. Liao and L. Shao  are with the Inception Institute of Artificial Intelligence, Abu Dhabi, United Arab Emirates. L. Shao is also with the Mohamed bin Zayed University of Artificial Intelligence, Abu Dhabi, United Arab Emirates.}
\thanks{Corresponding author: Fang Zhao (email: zhaofang0627@gmail.com).}}

%
%

\markboth{Journal of \LaTeX\ Class Files,~Vol.~14, No.~8, August~2015}%
{Shell \MakeLowercase{\textit{et al.}}: Bare Demo of IEEEtran.cls for IEEE Journals}
%



\maketitle

\begin{abstract}
Unsupervised domain adaptation (UDA) for person re-identification is challenging because of the huge gap between the source and target domain. A typical self-training method is to use pseudo-labels generated by clustering algorithms to iteratively optimize the model on the target domain. However, a drawback to this is that noisy pseudo-labels generally cause trouble in learning. To address this problem, a mutual learning method by dual networks has been developed to produce reliable soft labels. However, as the two neural networks gradually converge, their complementarity is weakened and they likely become biased towards the same kind of noise. This paper proposes a novel light-weight module, the Attentive WaveBlock (AWB), which can be integrated into the dual networks of mutual learning to enhance the complementarity and further depress noise in the pseudo-labels. Specifically, we first introduce a parameter-free module, the WaveBlock, which creates a difference between features learned by two networks by waving blocks of feature maps differently. Then, an attention mechanism is leveraged to enlarge the difference created and discover more complementary features. Furthermore, two kinds of combination strategies, \textit{i.e.} pre-attention and post-attention, are explored. Experiments demonstrate that the proposed method achieves state-of-the-art performance with significant improvements on multiple UDA person re-identification tasks. We also prove the generality of the proposed method by applying it to vehicle re-identification and image classification tasks. Our codes and models are available at: \href{https://github.com/WangWenhao0716/Attentive-WaveBlock}{AWB}.

\end{abstract}

%
\IEEEpeerreviewmaketitle

\section{Introduction}
%
%
%
%
\IEEEPARstart{T}{he} target of person re-identification (re-ID) is to match images of a person across different camera views. Because of its extensive numbers of applications, person re-ID has attracted attention from both academia and industry. In recent years, with the development of deep learning, supervised re-ID methods, such  as \cite{sun2018beyond, wang2018learning, quan2019auto, chen2019abd, luo2019strong, zhou2019omni, zheng2019joint, chen2019self}, have gained impressive progress. However, there still exist several drawbacks. First, these methods require intensive manual labeling, which is expensive and time-consuming. Second, due to the domain gap, there is a significant performance drop when a model trained on a source domain is tested on a target domain \cite{deng2018image,fan2018unsupervised}. Therefore, unsupervised domain adaptation (UDA) was introduced, which aims at learning a model on a labeled source domain and adapting it to an unlabeled target domain.\par 
Image-level adaptation, such as \cite{deng2018image,wei2018person}, uses a generative adversarial network (GAN) \cite{goodfellow2014generative} to transfer the image styles of the source domain to a target domain. Feature-level method like \cite{zhong2019invariance} investigates underlying feature invariance. However, the performances of these approaches are still unsatisfactory when compared to their fully-supervised counterparts. Recently, several clustering based methods, such as \cite{song2020unsupervised, zhang2019self, fu2019self, kumar2020unsupervised}, have been proposed, which employ clustering algorithms to group unannotated target images to generate pseudo-labels for training. Although they achieve state-of-the-art performance in various UDA tasks, their abilities are hindered by noisy pseudo-labels caused by the imperfect clustering algorithms and the limited feature transferability.\par
\begin{figure*}
\centering
\subfigure[Original image]{
\begin{minipage}[t]{0.22\linewidth}
\centering
\includegraphics[width=0.6in]{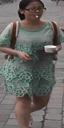}
\end{minipage}%
}%
\subfigure[MMT \cite{ge2020mutual}]{
\begin{minipage}[t]{0.23\linewidth}
\centering
\includegraphics[width=1.2in]{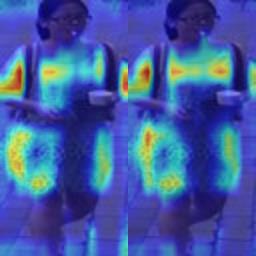}
\end{minipage}%
}
\subfigure[WaveBlock]{
\begin{minipage}[t]{0.23\linewidth}
\centering
\includegraphics[width=1.2in]{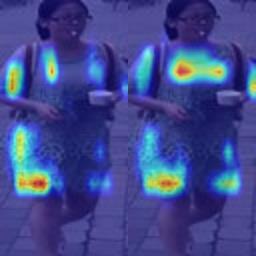}
\end{minipage}%
}
\subfigure[AWB]{
\begin{minipage}[t]{0.23\linewidth}
\centering
\includegraphics[width=1.2in]{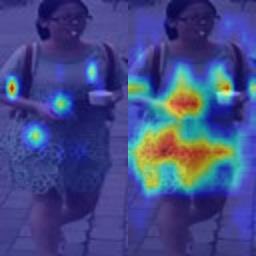}
\end{minipage}%
}
\centering
\caption{The gradient-weighted class activation maps of MMT \cite{ge2020mutual}, WaveBlock, and AWB. The differences in Frobenius norm between two maps for the three methods are $1.58$, $2.04$ and $4.83$, respectively. For MMT \cite{ge2020mutual}, the “focus” of the two networks is similar, and the activated areas are not very discriminative (not covering the full body and some are on the background). When using WaveBlock, the “focus” of the two networks becomes different. By combining attention modules with WaveBlock, the difference becomes even larger, and the activated areas of one network cover most of the body while the activated areas of another cover important points.}
\label{fm}
\end{figure*}

 
To address the aforementioned problem, a dual network framework, Mutual Mean-Teaching (MMT) \cite{ge2020mutual} was proposed, which trains two networks simultaneously and utilizes a temporally averaged model to produce reliable soft labels as supervision signals. Although this design reduces the amplification of training error to some degree, as the two networks converge, as shown in Fig. \ref{fm}, they unavoidably become more and more similar, which weakens their complementarity and may make them bias towards the same kind of noise. This limits further improvement in performance.\par 
To overcome the above limitations, we propose a novel module, namely the Attentive WaveBlock (AWB), under the dual network framework. The critical idea behind AWB is to create a difference between features learned by two neural networks to enhance their complementarity. In particular, we first introduce the WaveBlock to modulate feature maps of the two networks with different block-wise waves. Then, an attention mechanism is utilized to force the networks to focus on discriminative features in these regions, which further enlarges the difference between them. Here two kinds of combinations are designed, \textit{i.e.} pre-attention (Pre-A) and post-attention (Post-A), to produce such different and discriminative features. For Pre-A, the attention modules first learn discriminative features, and then WaveBlocks wave regions differently. For Post-A, WaveBlocks first generate different waves, and then the attention modules learn discriminative features on the different waves. In Fig. \ref{fm}, we visualize the feature attention maps of the three mutual learning methods using a gradient-weighted class activation map \cite{selvaraju2017grad} and compute the difference in Frobenius norm between two maps $A$ and $B$, which is ${\left\| {A - B} \right\|_F} = \sqrt {\sum\nolimits_{i,j} {{{\left| {{a_{ij}} - {b_{ij}}} \right|}^2}} } $. As shown in Fig. \ref{fm}, from MMT \cite{ge2020mutual} to WaveBlock, the difference increases to some degree. Further, from WaveBlock to AWB, the attention mechanism enlarges the difference created before.

Our contributions are summarized as follows:
\begin{itemize}
  \item We introduce a parameter-free module, the WaveBlock, that can create a difference between features learned by the dual network framework. It enhances the complementarity of the two networks and reduces the possibility that they become biased towards the same kind of noise.
  \item We propose to utilize an attention mechanism to enlarge the difference between networks on the basis of the WaveBlock and design two kinds of combination strategies, \textit{i.e.} pre-attention and post-attention. 
  \item The AWB module significantly improves performances on UDA tasks for person re-ID, with negligible computational increase. Compared with the state-of-the-art methods, we obtain improvements of $9.8\%$, $5.8\%$, $6.2\%$, $6.1\%$, $6.4\%$ and $10.0\%$ in mAP on Duke-to-Market, Market-to-Duke, Duke-to-MSMT, Market-to-MSMT, MSMT-to-Duke, and MSMT-to-Market re-ID tasks. 
\end{itemize}

\section{Related Works}
\subsection{General Domain Adaptation}
Domain adaptation aims to transfer the learned knowledge from a well-labeled source domain to a target domain.  Usually, the two domains have different data distribution, which is known as the domain gap and prevents performance improvement. Most domain adaptation algorithms \cite{long2016unsupervised, saito2018maximum, ganin2015unsupervised, Kang_2019_CVPR, sener2016learning, Pinheiro_2018_CVPR} can be categorized into two classes, i.e. \ feature-level and sample-level.  For instance,  MDD \cite{li2020maximum} minimizes the inter-domain divergence and maximizes the intra-class density. The method solves the problem from the feature level. From the sample level, SBADA-GAN \cite{russo2018source} introduces a symmetric mapping among domains to reconstruct source-like target images. Some recent researches argue that feature-level and sample-level are both critical for the unsupervised domain adaptation tasks.  Therefore,  \cite{li2019locality} proposes to jointly exploit feature adaptation with distribution matching and sample adaptation with landmark selection. The experimental results are quite promising. However, for many real-world applications, too much training data in the target domain is a burden. To address this, Faster Domain Adaptation Networks \cite{li2021faster} are proposed, which achieve comparable accuracy with much less computing resources.  However, the general domain adaptation pipeline, which shares the same classes between domains, is not suitable for person re-ID tasks because the identities in two re-ID domains are different. Therefore the design of re-ID-specific domain adaptation algorithms is essential.

\subsection{Unsupervised Domain Adaptation for Person Re-ID}
Mainstream algorithms for UDA tasks can be categorized into three classes. The first is image-level methods. They use a GAN to transfer the source domain images to the target-domain style \cite{ye2020deep}. For instance, PTGAN \cite{wei2018person} transfers knowledge, while SPGAN \cite{deng2018image} focuses on self-similarity and domain-dissimilarity. However, unfortunately, the performance of these methods lags far behind their fully-supervised counterparts. The second category is feature-level methods. For example, \cite{zhong2019invariance} investigates three types of underlying invariance, \textit{i.e.} exemplar-invariance, camera-invariance and neighborhood-invariance. The last category is clustering based adaptation. These methods \cite{fan2018unsupervised, lin2019bottom, zhang2019self,fu2019self} follow a similar general pipeline: they first pre-train on the source domain and then transfer the learned parameters to fit the target domain. Due to the imperfect clustering algorithms and big domain variance, the generated pseudo-labels tend to contain noise, which hinders further improvement in performance. Although, MMT \cite{ge2020mutual} was introduced to alleviate this problem by using a couple of neural networks to generate soft pseudo-labels, as the training process goes on, the two neural networks tend to converge and unavoidably share a high similarity. Therefore, it is necessary to consider how to create different networks and enhance the complementarity. This is the starting point of our AWB.

\subsection{Attention Mechanism}
Attention has been widely used to enhance representation learning in the fields of image classification \cite{wang2017residual,peng2017object,xiang2020semi}, object detection \cite{chen2018reverse,zhang2018progressive,fan2019shifting} and so on. For instance, Squeeze-and-Excitation (SE) block \cite{hu2018squeeze} recalibrates
channel-wise feature responses and convolutional block attention module (CBAM) \cite{woo2018cbam} further uses channel attention and spatial attention to explore ``what" and ``where" to focus. By stacking attention modules which can generate       module-adaptation and attention-aware features, Residual Attention Network \cite{wang2017residual} is built. Non-local block \cite{wang2018non} explores the relationship between different positions on feature maps and exploits global features. In the person re-ID community, fully-supervised state-of-the-arts algorithms, such as ConsAtt \cite{zhou2019discriminative}, SCAL \cite{chen2019self}, SONA \cite{xia2019second}, and ABD-Net \cite{chen2019abd}, on several datasets (Market-1501 \cite{zheng2015scalable}, DukeMTMC \cite{zheng2017unlabeled}, CUHK03 \cite{li2014deepreid}, MSMT17 \cite{wei2018person}) adopt an attention scheme. However, nearly all aforementioned works utilize attention mechanism to discover discriminative or critical features to boost the performance. In our work, beyond the stated functions, we find attention mechanism can enlarge the difference created by WaveBlocks. Therefore, by integrating attention mechanism, the improved performance comes from more complementary and more discriminative features extracted by two neural networks.


\subsection{Drop-series}
Dropout \cite{JMLR:v15:srivastava14a} was proposed as a regularization method to prevent overfitting problem by dropping units in fully connected layers. Instead of dropping discrete units, DropBlock \cite{ghiasi2018dropblock} drops units in a contiguous region of a feature map. Batch DropBlock Network (BDB) \cite{dai2019batch} uses a global branch and a feature dropping branch to keep the global salient representations and reinforce the attentive feature learning of local regions. Wu \cite{wu2020diversity} uses multiple dropping branches on the basis of BDB \cite{dai2019batch} to further boost the performance. Different from Dropout \cite{JMLR:v15:srivastava14a}, the proposed WaveBlock modulates a continuous region of a feature map like DropBlock \cite{ghiasi2018dropblock}. However, unlike DropBlock \cite{ghiasi2018dropblock} which may drop some discriminant information randomly, the proposed WaveBlock modulates a given feature map with different waves. This design preserves the original information to some degree. Comparing with BDB \cite{dai2019batch}, which increases the computing burden by introducing another branch, the proposed WaveBlock is totally parameter-free.

\section{Proposed Method}
In this section, we first simply review the Mutual Mean-Teaching (MMT) framework, then introduce the proposed WaveBlock module. Finally, two different strategies for combining attention mechanism with WaveBlock are presented.

\subsection{MMT framework Revisit}
Briefly, the MMT framework includes two identical networks with different initializations. Its pipeline is as follows: first, the two networks are pre-trained on the source domain to obtain initialized parameters. Then, in each epoch, offline hard pseudo-labels are generated using a clustering algorithm. In each iteration of a given epoch, refined soft pseudo-labels are produced by the two networks. The hard pseudo-labels and refined soft pseudo-labels generated by one network are then used together to supervise the learning process of the other network. Finally, again in each iteration, the temporally averaged models are updated and used for prediction. For more details, please refer to \cite{ge2020mutual}.
\begin{figure}
\centering
\includegraphics[width=6cm]{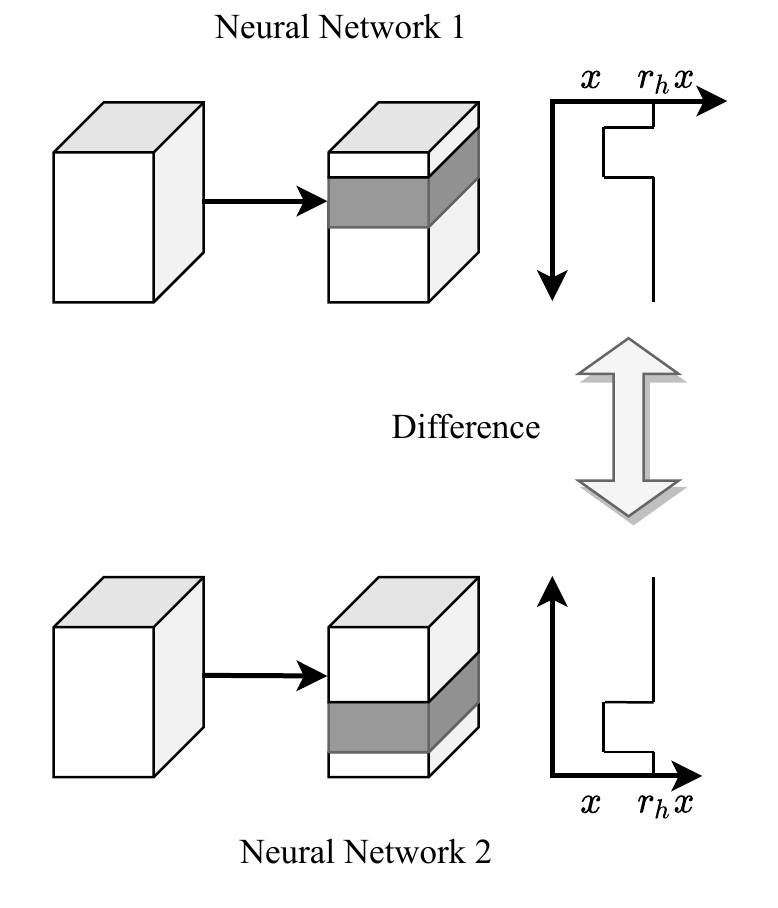}
\caption{Overview of the WaveBlock module, which creates a difference between features learned by two networks by waving blocks of feature maps differently. Specifically, a block is randomly selected and kept the same, while feature values of other blocks are multiplied $r_h$ times to form a wave.}
\label{WB}
\end{figure}
\subsection{WaveBlock}
In order to enhance the complementarity of the two networks, we first introduce the WaveBlock module to create a difference  between features learned by the networks, which is illustrated in Fig. \ref{WB}. Instead of dropping blocks as in \cite{ghiasi2018dropblock} which may lose discriminant information, WaveBlocks modulate a given feature map with different block-wise waves, so that differences are created between dual networks, and meanwhile the original information is preserved to some extent.\par 

Given a feature map $F \in {R^{C \times H \times W}}$, where $C$ is the number of channels, $H$ and $W$ are spatial height and width, respectively, a waving width rate $r_w$, and a waving height rate $r_h$, we first generate a random integer with uniform distribution: 
\begin{equation}
X \sim U(0,[H\cdot(1-r_w)]),
\end{equation}
where $[ \cdot ]$ is the rounding function. Then, the WaveBlock modulated feature map is defined as ${F^ *} \in {R^{ C \times H \times W}}$:
\begin{equation}
F_{ijk}^* = \left\{ {\begin{array}{*{20}{l}}
{{F_{ijk}},X \le j < X + \left[ {H \cdot r_w} \right],}\\
{r_h \cdot {F_{ijk}},otherwise.}
\end{array}} \right.
\end{equation}
where $i$, $j$, and $k$ respectively represent the coordinates of the dimension, height, and width of the feature map. This design modulates a given feature map with block-wise waves and meanwhile original information is kept to some degree. When applying WaveBlocks to the feature maps $F_1,F_2$ of two networks, respectively, the difference between the networks can be created by waving differently on blocks of feature maps. Let $F_1^*$, $F_2^*$ denote the output feature maps of WaveBlock and $X_1$, $X_2$ indicate the waving random integers generated on the two networks; we will calculate the probability that the same wave is generated for both. For simplicity, it is assumed that $F_1$ and $F_2$ have the same size.\par 
In order to enable $F_1^ *  = F_2^ * $, we should make $X_1 = X_2$. 
Since
\begin{equation}
P\left( {X_1 = X_2} \right) = \frac{{\left[ {H \cdot \left( {1 - r_w} \right)} \right]}}{{{{\left[ {H \cdot \left( {1 - r_w} \right)} \right]}^2}}} = \frac{1}{{\left[ {H \cdot \left( {1 - r_w} \right)} \right]}},
\end{equation}
we have
\begin{equation}
P\left( F_1^ *  = F_2^ * \right) = P\left( {{X_1} = {X_2}} \right) = \frac{1}{{\left[ {H \cdot \left( {1 - r_w} \right)} \right]}}.
\end{equation}
If multiple GPUs are used for training, $X$ will be generated independently in each GPU. In practice, $r_w$ is set as $0.3$ experimentally and four GPUs are used. Then, on feature maps with $H=32$, we have
\begin{equation}
P\left( {F_1^* = F_2^*} \right) = \frac{1}{{{{\left[ {H \cdot \left( {1 - r_w} \right)} \right]}^4}}} = 4.27 \cdot {10^{-6}}.
\end{equation}
Because the probability is too small for the waves of the two networks to be the same, we may say that there is always a difference created between them.

\subsection{Attentive WaveBlock}
\begin{figure}[t]  
\centering  
\includegraphics[width=0.5\textwidth]{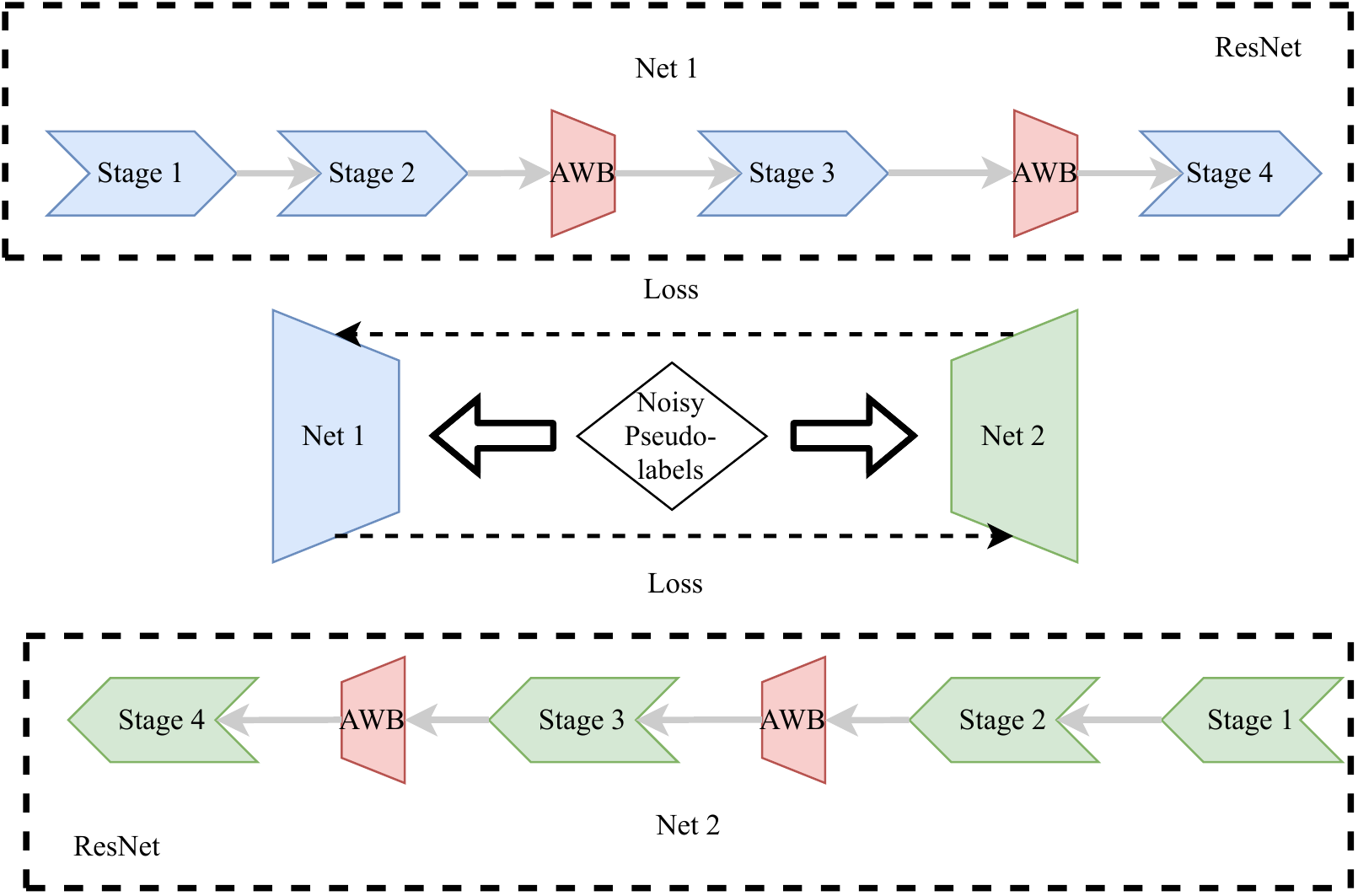} 
\caption{The overview of our complementarity-enhanced mutual networks. Through the proposed AWB modules, the two networks learn different and discriminative features. The noise in pseudo-labels is suppressed to some degree.}  
\label{overview}  
\end{figure}

To enlarge the difference created by WaveBlocks and find more discriminative and more complementary features, the attention mechanism is integrated with the WaveBlock module in this section. Two kinds of combination strategies are designed, including pre-attention (Pre-A) and post-attention (Post-A).  Note that the attention modules used in the two networks do not share weights. The overview of MMT \cite{ge2020mutual} integrated with AWB is shown in Fig. \ref{overview}.

\subsubsection{Attention Mechanism}
To show that the proposed WaveBlock can be combined to general attention methods, two kinds of attention mechanisms are tried here. The first one is the convolutional block attention module (CBAM) \cite{woo2018cbam}. Given a feature map $F \in {R^{C \times H \times W}}$, CBAM exerts a channel attention map ${M_c} $ and a spatial attention map ${M_s}$ on $F$ sequentially:
\begin{equation}
{K_1} = {M_c}\left( F \right) \otimes F,
\end{equation}
\begin{equation}
{K_2} = {M_s}\left( {{K_1}} \right) \otimes {K_1},
\end{equation}
where $ \otimes $ denotes element-wise multiplication. In CBAM, the channel attention exploits the inter-channel relationship of features, while the spatial attention focuses on ``where" an informative part is located.  \par 
In the original paper of CBAM \cite{woo2018cbam}, CBAM is integrated into a block of ResNet \cite{he2016deep}. However, due to the depth of ResNet \cite{he2016deep}, the computing burden increases to some degree. Therefore, we improve the original CBAM to arrange it between sequential stages of ResNet \cite{he2016deep}. In each improved CBAM module, the original feature map $F$ is added to the modified one $K_2$ to obtain the final one $K_3$, which aims to avoid the information loss. \par 
The second attention mechanism is the Non-local block \cite{wang2018non}. Here, its simplified version is adopted. Let $F \in {R^{ C \times H \times W}}$ denote a feature map for Non-local block and $\theta$ denote a $1 \times 1$ convolution. Through $\theta$, the number of channels of $F$ are reduced from $C$ to $C/2$, \textit{i.e.} $\theta \left( {{F}} \right) \in {R^{ \frac{C}{2} \times H \times W}}$. Similarly, another $1 \times 1$ convolution $\phi$ also reduces the number of channels from $C$ to $C/2$, \textit{i.e.} $\phi \left( {{F}} \right) \in {R^{ \frac{C}{2} \times H \times W}}$. Then we collapse the spatial dimension of $\theta \left( {{F}} \right)$ and $\phi \left( {{F}} \right)$ into a single dimension, \textit{i.e.} $\theta '\left( {{F}} \right) \in {R^{ \frac{C}{2} \times HW}}$, $\phi' \left( {{F}} \right) \in {R^{ \frac{C}{2} \times HW}}$. We obtain our matrix $J \in {R^{HW \times HW}}$:
\begin{equation}
J = {\left( {\theta '\left( F \right)} \right)^T} \cdot \phi '\left( F \right).
\end{equation}
Next, we adopt $\frac{1}{{H \times W}}$ as the scaling factor for $J$, without using $softmax$. In the other branch, $F$ is fed into a function $g$, which is a $1 \times 1$ convolution followed by a batch normalization layer. Similarly, we collapse the spatial dimension of $g({F})$ into a single dimension and further apply a transpose to get $g'({F}) \in {R^{ HW \times \frac{C}{2}}}$. Finally, we multiply $J$ with $g'({F})$, transpose and reshape its dimensions to $ \frac{C}{2} \times H \times W$, and use another $1 \times 1$ convolution $h$ to restore the channel dimension to $C$. The result is denoted as $I$. Also, the final feature map is obtained by the sum of $I$ and $F$.\par 
\begin{figure*}[t]
\centering    
 
\subfigure[Pre-attention] 
{	
	\begin{minipage}{13cm}
	\centering
	\includegraphics[scale=0.65]{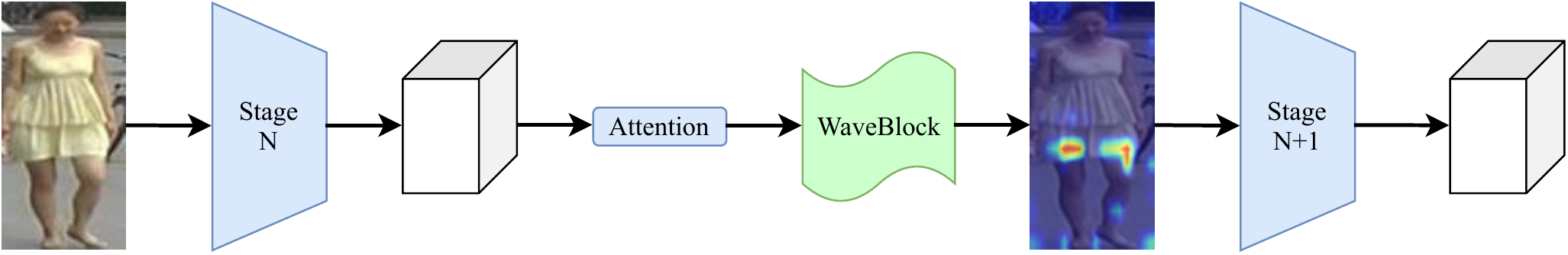}  
	\label{Pre-A} 
	\end{minipage}
}

\subfigure[Post-attention] 
{
	\begin{minipage}{13cm}
	\centering
	\includegraphics[scale=0.65]{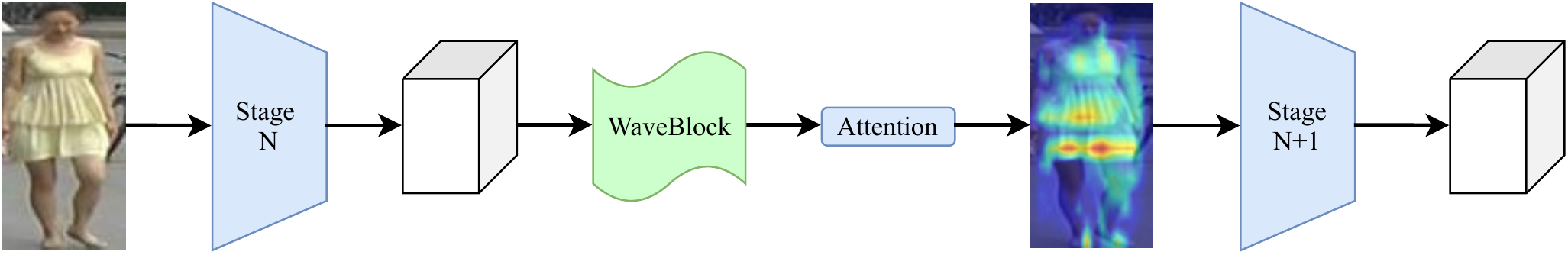}   
	\label{Post-A} 
	\end{minipage}
}
 
\caption{Two different combination strategies for the attention module and WaveBlock. The benefit of Pre-A is the attention modules can be calculated using complete features. The advantage of Post-A is that directly applying attention module on waved features is more efficient to enlarge the difference created. It should be noted that the use of Pre-A and Post-A is separated, \textit{i.e.} when we are under the Pre-A architecture, the two networks in the framework only use Pre-A with WaveBlock, and that is the same for Post-A.}
 
\end{figure*}

\subsubsection{Pre-Attention}
As illustrated in Fig. \ref{Pre-A}, to combine the attention module with the WaveBlock, we first try to arrange it before the WaveBlock, which is called the Pre-attention (Pre-A) strategy. In this way, the attention modules first learn discriminative features, and then WaveBlocks wave regions differently to produce different and discriminative features. Given a feature map $F \in {R^{ C \times H \times W}}$, WaveBlock is applied to either of the two attention modules mentioned before and obtain:\par

\begin{equation} 
{F^*} = WaveBlock\left( {{M_s}\left( {{M_c}\left( F \right) \otimes F} \right) \otimes \left( {{M_c}\left( F \right) \otimes F} \right) + F} \right),
\end{equation}

or
\begin{equation} 
F^* = WaveBlock\left( {h\left( {{{\left( {\theta '\left( F \right)} \right)}^T} \cdot \phi '\left( F \right) \cdot g'\left( F \right)} \right) + F} \right).
\end{equation}
Here, the attention modules are used to enlarge the difference of the backward gradients generated by the WaveBlock. That means the updating process of the two attention modules’ weights uses different gradients modulated by the WaveBlock. When the gradients are changed, the focus of attention module will also be changed. The changes of two phases lead to a larger difference. For instance, although the WaveBlock is able to make the two networks work on different regions of feature maps, some features learned from non-discriminative regions, such as backgrounds, may still be similar. By combining the attention modules with the WaveBlock, the two networks focus on different and discriminative regions, such as the human body, and thus can learn more different features. 
The advantage of Pre-A is that the attention weights can be computed by using the complete feature maps. This is more beneficial to CBAM because the convolution used to compute its spatial attention will not be affected near the border of waved regions. 

\subsubsection{Post-Attention}
The second combination strategy is shown in Fig. \ref{Post-A}. We arrange the attention mechanism after the WaveBlock, which is named as post-attention (Post-A). Correspondingly, the WaveBlocks first wave regions differently, and then the attention modules learn discriminative features on the waved regions to produce different and discriminative features. Given a feature map $F \in {R^{ C \times H \times W}}$, after passing through the WaveBlock, either of the two attention modules mentioned before can be applied. This produces:
\begin{equation} 
\widetilde F = WaveBlock\left( F \right),
\end{equation}
\begin{equation} 
{F^ * } = {M_s}\left( {{M_c}\left( {\widetilde F} \right) \otimes \widetilde F} \right) \otimes \left( {{M_c}\left( {\widetilde F} \right) \otimes \widetilde F} \right) + {\widetilde F},
\end{equation}
or

\begin{equation}
F^* = h\left( {{{\left( {\theta '\left( {{\widetilde F}} \right)} \right)}^T} \cdot \phi '\left( {{\widetilde F}} \right) \cdot g'\left( {{\widetilde F}} \right)} \right) + {\widetilde F}.
\end{equation}
 
Compared with Pre-A, although the waved regions may affect the computation of the attention weights, directly applying the attention modules on the different waved regions is more efficient for enlarging different features.  An understanding for enlarging difference from the prospective of gradient-weight class activation map is shown as follows. An attention mechanism can be understood as a $0 \sim 1$ mask. Given two maps $X$ and $Y$, as defined before, the difference is quantified as  ${\left\| {X - Y} \right\|_F} = \sqrt {\sum\nolimits_{i,j} {{{\left| {{x_{ij}} - {y_{ij}}} \right|}^2}} } $.  Using an attention mechanism, $X$ is turned to ${\widetilde X} = Attention(X)+X$. Similarly, ${\widetilde Y} = Attention(Y)+Y$. Then, assuming the attention mask is the same for the two maps, the new difference is calculated as: 
\begin{equation}
\begin{array}{l}
{\left\| {\tilde X - \tilde Y} \right\|_F}\\
 = {\left\| {Attention\left( X \right) - Attention\left( Y \right) + X - Y} \right\|_F}\\
 = \sqrt {\sum\nolimits_{i,j} {\left( {{\alpha _{ij}} + 1} \right){{\left| {{x_{ij}} - {y_{ij}}} \right|}^2}} } \\
 \ge \sqrt {\sum\nolimits_{i,j} {{{\left| {{x_{ij}} - {y_{ij}}} \right|}^2}} } = {\left\| {X - Y} \right\|_F}
\end{array}
\end{equation}
where: $\alpha _{ij}$ is a weight from $0 \sim 1$ attention mask. This formula implies that, after applying attention mechanism to two gradient-weight class activation maps, the quantified difference is enlarged.  Moreover, when the attention mask is different, the quantified difference can be enlarged much more.
Post-A is more beneficial to the Non-local block because the non-local operation reduces the impact of waved regions.

\begin{table*}
\caption{Comparison between the proposed method and state-of-the-art algorithms. The results are reported on Market-1501 \cite{zheng2015scalable}, DukeMTMC \cite{zheng2017unlabeled} and MSMT17 \cite{wei2018person}. (*) implies the implementation is based on the codes provided by the original paper.}
\label{SOTA}
\vspace*{2mm}
\small
  \begin{tabularx}{\hsize}{p{4.5cm}|YYYY|YYYY}
    \hline
    \multicolumn{1}{c|}{\multirow{2}{*}{Methods}} &
    \multicolumn{4}{c|}{Duke-to-Market} &
    \multicolumn{4}{c}{Market-to-Duke} \\
    \cline{2-9}
      & mAP & rank-$1$ & rank-$5$ & rank-$10$ & mAP & rank-$1$ & rank-$5$ & rank-$10$ \\
    \hline\hline
    PUL \cite{fan2018unsupervised} & $20.5$ & $45.5$ & $60.7$ & $66.7$ & $16.4$ &$30.0$ &$43.4$&$48.5$\\
SPGAN \cite{deng2018image} & $22.8$ & $51.5$ & $70.1$ & $76.8$ & $22.3$ &$41.1$ &$56.6$&$63.0$\\
TJ-AIDL \cite{wang2018transferable} & $26.5$ & $58.2$ & $74.8$ & $81.1$ & $23.0$ &$44.3$ &$59.6$&$65.0$\\
CFSM \cite{chang2019disjoint} & $28.3$ & $61.2$ & $-$ & $-$ & $27.3$ &$49.8$ &$-$&$-$\\
UCDA \cite{qi2019novel} & $30.9$ & $60.4$ & $-$ & $-$ & $31.0$ &$47.7$ &$-$&$-$\\
HHL \cite{zhong2018generalizing} & $31.4$ & $62.2$ & $78.8$ & $84.0$ & $27.2$ &$46.9$ &$61.0$&$66.7$\\
BUC \cite{lin2019bottom} & $38.3$ & $66.2$ & $79.6$ & $84.5$ & $27.5$ &$47.4$ &$62.6$&$68.4$\\
ARN \cite{li2018adaptation} & $39.4$ & $70.3$ & $80.4$ & $86.3$ & $33.4$ &$60.2$ &$73.9$&$79.5$\\
CDS \cite{wu2019clustering} & $39.9$ & $71.6$ & $81.2$ & $84.7$ & $42.7$ &$67.2$ &$75.9$&$79.4$\\
ECN \cite{zhong2019invariance} & $43.0$ & $75.1$ & $87.6$ & $91.6$ & $40.4$ &$63.3$ &$75.8$&$80.4$\\
PDA-Net \cite{li2019cross} & $47.6$ & $75.2$ & $86.3$ & $90.2$ & $45.1$ &$63.2$ &$77.0$&$82.5$\\
UDAP \cite{song2020unsupervised}  & $53.7$ &$75.8$ &$89.5$&$93.2$ & $49.0$ & $68.4$ & $80.1$ & $83.5$\\
CR-GAN \cite{chen2019instance}  & $54.0$ &$77.7$ &$89.7$&$92.7$ & $48.6$ & $68.9$ & $80.2$ & $84.7$\\
PCB-PAST \cite{zhang2019self} & $54.6$ & $78.4$ & $-$ & $-$ & $54.3$ &$72.4$ &$-$&$-$\\
AE \cite{ding2020adaptive} & $58.0$ & $81.6$ & $91.9$ & $94.6$ & $46.7$ &$67.9$ &$79.2$&$83.6$\\
SSG \cite{fu2019self} & $58.3$ & $80.0$ & $90.0$ & $92.4$ & $53.4$ &$73.0$ &$80.6$&$83.2$\\
pMR-SADA\cite{wang2020smoothing} & $59.8$ & $83.0$ & $91.8$ & $94.1$ & $55.8$ &$74.5$ &$85.3$&$88.7$\\
MMCL \cite{wang2020unsupervised} & $60.4$ & $84.4$ & $92.8$ & $95.0$ & $51.4$ &$72.4$ &$82.9$&$85.0$\\
ACT \cite{yang2019asymmetric} & $60.6$ & $80.5$ & $-$ & $-$ & $54.5$ &$72.4$ &$-$&$-$\\
SNR \cite{jin2020style} & $61.7$ & $82.8$ & $-$ & $-$ & $58.1$ &$76.3$ &$-$&$-$\\
ECN++ \cite{zhong2020learning} & $63.8$ & $84.1$ & $92.8$ & $95.4$ & $54.4$ &$74.0$ &$83.7$&$87.4$\\
AD-cluster \cite{zhai2020ad} & $68.3$ & $86.7$ & $94.4$ & $96.5$ & $54.1$ &$72.6$ &$82.5$&$85.5$\\
MMT \cite{ge2020mutual} & $71.2$ & $87.7$ & $94.9$ & $96.9$ & $65.1$ &$78.0$ &$88.8$&$92.5$\\
\hline
WaveBlock& $76.3$ & $90.9$ & $96.7$ & $97.7$ & $68.6$ &$82.4$ &$91.4$&$94.0 $\\
I-CBAM& $75.4$ & $90.1$ & $96.1$ & $97.4$ & $67.8$ &$80.2$ &$90.3$&$93.3$\\
Non-local& $79.0$ & $92.5$ & $97.2$ & $98.5$ & $69.6$ &$82.4$ &$91.0$&$93.8$\\
AWB (Pre-A with I-CBAM)& $78.8$ & $92.2$ & $97.1$ & $98.1$ & $70.0$ &$82.9$ &$91.4$&$ 93.9$\\
AWB (Post-A with Non-local)& $\textbf{81.0}$ & $\textbf{93.5}$ & $\textbf{97.4}$ & $\textbf{98.3}$ & $\textbf{70.9}$ &$\textbf{83.8} $ &$\textbf{92.3}$&$\textbf{94.0}$\\
    \hline
    \hline
    \multicolumn{1}{c|}{\multirow{2}{*}{Methods}} &
    \multicolumn{4}{c|}{Duke-to-MSMT} &
    \multicolumn{4}{c}{Market-to-MSMT} \\
    \cline{2-9}
      & mAP & rank-$1$ & rank-$5$ & rank-$10$ & mAP & rank-$1$ & rank-$5$ & rank-$10$ \\
    \hline\hline
    RPTGAN \cite{wei2018person} & $3.3$ & $11.8$ & $-$ & $27.4$ &  $2.9$ &$10.2$ &$-$&$24.4$\\
ECN \cite{zhong2019invariance} & $10.2$ &$30.2$ &$41.5$&$46.8$ & $8.5$ & $25.3$ & $36.3$ & $42.1$ \\
AE \cite{ding2020adaptive} & $11.7$ & $32.3$ & $44.4$ & $50.1$ & $9.2$ &$25.5$ &$37.3$&$42.6$\\
SSG \cite{fu2019self}  & $13.3$ &$32.2$ &$-$&$51.2$ & $13.2$ & $31.6$ & $-$ & $49.6$\\
ECN++ \cite{zhong2020learning} & $16.0$ & $42.5$ & $55.9$ & $61.5$ & $15.2$ &$40.4$ &$53.1$&$58.7$\\
MMCL \cite{wang2020unsupervised} & $16.2$ & $43.6$ & $54.3$ & $58.9$ & $15.1$ &$40.8$ &$51.8$&$56.7$\\
MMT \cite{ge2020mutual}  & $23.3$ &$50.1$ &$63.9$&$69.8$ & $22.9$ & $49.2$ & $63.1$ & $68.8$\\
\hline
WaveBlock& $28.1$ & $58.5$ & $71.5$ & $76.4$ & $27.2$ &$56.1$ &$69.6$&$74.9$\\
I-CBAM& $24.1$ & $51.7$ & $65.5$ & $71.1$ & $24.2$ &$51.3$ &$65.1$&$70.8 $\\
Non-local& $28.4$ & $57.2$ & $70.7$ & $76.0$ & $28.5$ &$56.8$ &$\textbf{70.7}$&$ 75.7$\\
AWB (Pre-A with I-CBAM)& $\textbf{29.5}$ &$\textbf{61.0}$ &$
\textbf{73.5}$&$\textbf{77.9}$ & $27.3$ & $\textbf{57.8}$ & $\textbf{70.7}$ & $75.7$ \\
AWB (Post-A with Non-local)& $29.0$ & $57.9$ & $71.5$ & $76.6$ & $\textbf{29.0}$ & $57.3 $ & $\textbf{70.7} $ & $\textbf{75.9} $\\
    \hline
    \hline
    \multicolumn{1}{c|}{\multirow{2}{*}{Methods}} &
    \multicolumn{4}{c|}{MSMT-to-Duke} &
    \multicolumn{4}{c}{MSMT-to-Market} \\
    \cline{2-9}
      & mAP & rank-$1$ & rank-$5$ & rank-$10$ & mAP & rank-$1$ & rank-$5$ & rank-$10$ \\
    \hline\hline
PAUL \cite{Yang_2019_CVPR}  & $53.2$ & $72.0$ & $82.7$ & $86.0$ & $40.1$ &$68.5$ &$82.4$&$87.4$\\
MMT* \cite{ge2020mutual}  & $63.2$ &$76.5$ &$88.0$&$91.4$ & $69.4$ & $85.9$ & $94.4$ & $96.2$\\
\hline
WaveBlock& $67.1$ & $80.9$ & $89.9$ & $92.9$ & $75.8$ &$90.6$ &$96.4$&$97.9 $\\
I-CBAM& $66.4$ & $79.2$ & $89.7$ & $92.9$ & $71.6$ &$87.5$ &$94.9$&$ 97.1$\\
Non-local& $68.0$ & $81.4$ & $90.5$ & $93.1$ & $77.1$ &$90.4$ &$96.6$&$ 97.9$\\
AWB (Pre-A with I-CBAM)& $68.6$ &$\textbf{82.8}$ &$
\textbf{91.4}$&$93.1$ & $77.1$ & $91.2$ & $96.7$ & $97.8$ \\
AWB (Post-A with Non-local)& $\textbf{69.6}$ & $81.7$ & $90.5$ & $\textbf{93.4}$ & $\textbf{79.4}$ & $\textbf{92.6}$ & $\textbf{97.1} $ & $\textbf{98.2} $\\
    \hline

  \end{tabularx}
  \\
\end{table*}
\begin{table*}
\caption{Comparison between the proposed method and MMT \cite{ge2020mutual}. The results are reported on VehicleID \cite{liu_2016deep}, VeRi-776 \cite{liu2016deep}, and VehicleX \cite{naphade20204th}. (*) implies the implementation is based on the authors' codes.}
\label{SOTA_ve}
\vspace*{2mm}
\small
  \begin{tabularx}{\hsize}{p{4.5cm}|YYYY|YYYY}
    \hline
    \multicolumn{1}{c|}{\multirow{2}{*}{Methods}} &
    \multicolumn{4}{c|}{VehicleID-to-VeRi-776} &
    \multicolumn{4}{c}{VehicleX-to-VeRi-776} \\
    \cline{2-9}
      & mAP & rank-$1$ & rank-$5$ & rank-$10$ & mAP & rank-$1$ & rank-$5$ & rank-$10$ \\
    \hline\hline
  UDAP \cite{song2020unsupervised} & $35.8$ &$76.9$ &$85.8$&$89.0$ & $-$ &$-$ &$-$&$-$\\
MMT* \cite{ge2020mutual} & $36.1$ &$76.3$ &$82.8$&$87.4$ & $35.2$ &$75.9$ &$82.2$&$86.9$\\
\hline
WaveBlock& $36.9$ & $77.2$ & $83.9$ & $87.2$ & $36.0$ &$77.4$ &$84.3$&$ 88.3$\\
I-CBAM& $36.7$ & $77.2$ & $83.5$ & $88.0$ & $36.0$ &$77.1$ &$83.3$&$87.1$\\
Non-local& $36.9$ & $77.7$ & $83.6$ & $86.9$ & $37.1$ &$79.1$ &$84.6$&$89.0$\\
AWB (Pre-A with I-CBAM)& $\textbf{37.1}$ & $\textbf{79.1}$ & $\textbf{84.6}$ & $\textbf{88.5}$ & $36.7$ &$80.6$ &$86.0$&$89.7$\\
AWB (Post-A with Non-local)& $36.9$ & $78.2$ & $84.9$ & $88.3$ & $\textbf{37.2}$ &$\textbf{79.9}$ &$\textbf{85.2}$&$\textbf{89.2}$\\
    \hline
    
  \end{tabularx}
  \\
\end{table*}
\begin{table}
\centering
\caption{Comparison between the proposed method and DML \cite{zhang2018deep} for image classification task on ImageNet \cite{deng2009imagenet}. }
\label{cl}
\begin{tabular}{|c|c|c|c|c|} 
\hline  
\makecell[c]{Method}&\makecell[c]{DML \cite{zhang2018deep}}&\makecell[c]{WaveBlock}&\makecell[c]{AWB\\(I-CBAM)}&\makecell[c]{AWB\\(Non-local)}\\
\hline   
Top-$1$&$63.45\%$&$64.00\%$&$64.24\%$&$64.34\%$\\
\hline 
\end{tabular}
\end{table}

\section{Experiment}
\subsection{Datasets and Metrics}
\subsubsection{Person re-ID datasets}
\textbf{Market-1501} \cite{zheng2015scalable} is obtained using six different cameras. The dataset has $1,501$ labeled persons in $32,668$ images. For training, there are $12,936$ images of $751$ identities. For testing, the query has $3,368$ images and gallery has $19,732$ images. \textbf{DukeMTMC-reID} \cite{zheng2017unlabeled} contains $1,404$ persons from eight cameras. Among them, $16,522$ images of $702$ identities are used for training. For testing, there are $2,228$ queries, and $17,661$ gallery images. \textbf{MSMT17} \cite{wei2018person} is the most challenging and largest re-ID dataset. It consists of $126,441$ bounding boxes of $4,101$ identities taken by $15$ cameras. There are $32,621$ images for training while the query has $11,659$ images and the gallery has $82,161$ images. 
\subsubsection{Vehicle re-ID datasets}
To prove the generality of the proposed method, we also accomplish unsupervised domain adaptation task on three vehicle re-ID datasets, including Veri-776  \cite{liu2016deep}, VehicleID \cite{liu_2016deep}, and VehicleX \cite{naphade20204th}. \textbf{Veri-776} \cite{liu2016deep} is collected using $20$ different cameras. Among them, $37,746$ images of $575$ identities are used to train. The query set has $1,678$ images while the gallery set has $11579$ images. \textbf{VehicleID} \cite{liu_2016deep} contains $221,736$ vehicles. For training, $113,346$ images of $13,164$ identities are used. For testing, there are $5,693$ query images and $102,724$ gallery images. \textbf{VehicleX} \cite{naphade20204th} is a synthetic dataset generated by Unity engine \cite{yao2019simulating,tang2019pamtri} and translated to real-world style by SPGAN \cite{deng2018image}. The dataset has $192,150$ images of $1,362$ identities for training, and it does not have test part.
\begin{figure*}

\centering

\subfigure{
\centering
\begin{minipage}{1\textwidth}

\includegraphics[width=1\textwidth]{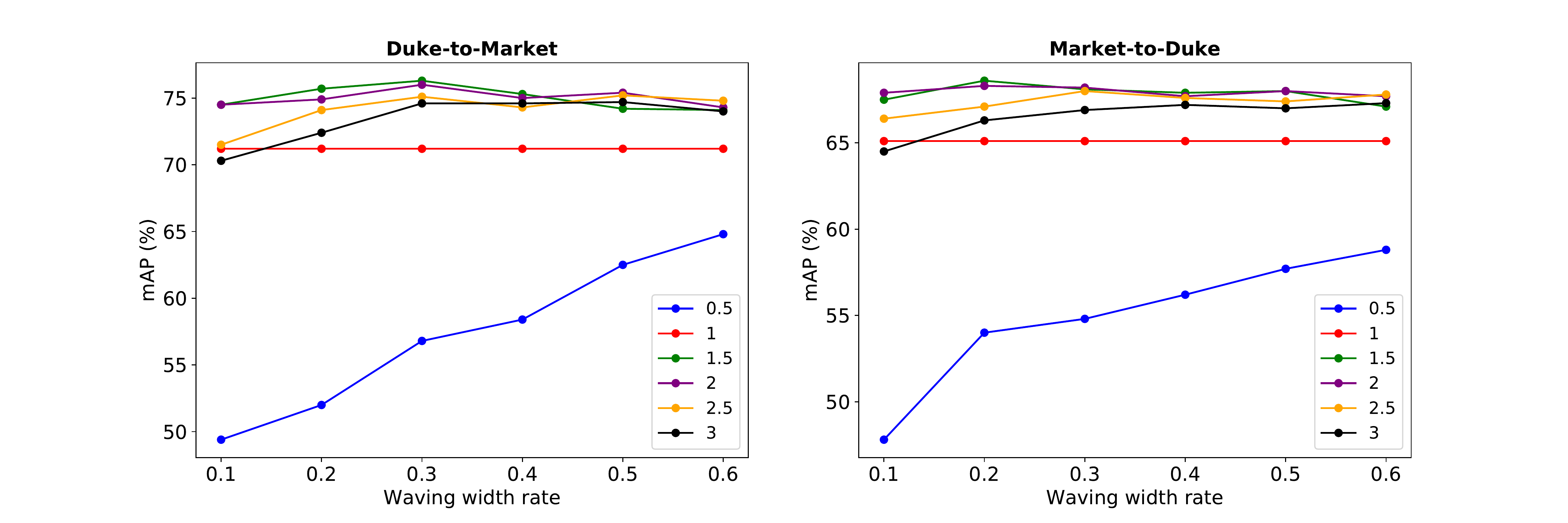}

\end{minipage}

}

\subfigure{
\centering
\begin{minipage}{1\textwidth}

\includegraphics[width=1\textwidth]{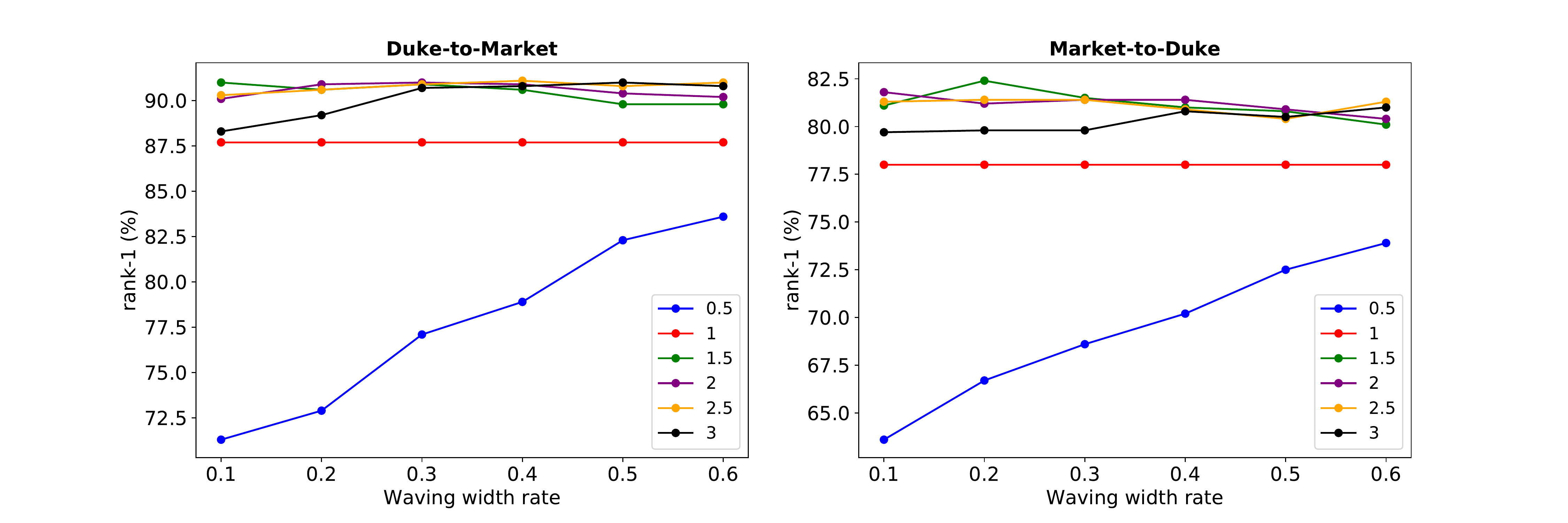}

\end{minipage}

}

 \caption{The mAP and rank-$1$ improvement under different experiment settings. Different lines represent different waving height rates. When waving height rate is larger than $1$,  WaveBocks improve the performance continuously.} \label{para}

\end{figure*}
\subsubsection{Image classification dataset}
To further verify the efficacy of the proposed method on large scale datasets, the proposed method is also applied to the classification task on ImageNet \cite{deng2009imagenet}. The dataset contains $1,000$ object classes with about $1.2$ million images for training and $50,000$ images for validation.
\subsubsection{Evaluation protocol} For re-ID datasets, to evaluate our algorithm, we adopt the mean average precision (mAP) and cumulative matching characteristic (CMC) at rank-1, rank-5, and rank-10. No post-processing, such as re-ranking \cite{zhong2017re}, is used and we utilize single-query evaluation protocols. For image classification dataset, top-1 classification accuracy is used as evaluation metric. 
\subsection{Experimental Settings}
We follow the same training settings as MMT \cite{ge2020mutual}, i.e. we do not adjust any hyper-parameters in MMT \cite{ge2020mutual} framework, and we just follow the same network structure (ResNet-50 \cite{he2016deep}) with MMT \cite{ge2020mutual} framework. Further, for the source-domain pre-training, to ensure that the improvement comes from a different mutual training but not an enhanced pre-trained network, no change is made to the pre-training process in MMT \cite{ge2020mutual} framework. \par

For the first stage of target-domain training, attention modules are trained without WaveBlock engaged. Specifically, two attention modules are plugged after Stage $2$ and Stage $3$ of the ResNet-50 \cite{he2016deep} backbone with random initialization. The two modules are trained for $10$ epochs with other parameters frozen. For the second stage target-domain training, WaveBlocks are added into two networks. Specifically, the attention modules are integrated with WaveBlocks after Stage $2$ and Stage $3$ of the ResNet-50 \cite{he2016deep} backbone to form AWB. For CBAM, the Pre-A design is used and for Non-local, the Post-A design is utilized. Because we successfully enhance the complementarity and make it some more difficult for the two neural networks biased towards the same kind of noise, the training process can last for more epochs. We train for $80$ epochs with all parameters engaged. When clustering, we select the optimal $k$ values of $k$-means following \cite{ge2020mutual}, \textit{i.e.} $500$ for Duke-to-Market, $700$ for Market-to-Duke, $1500$ for Duke-to-MSMT and Market-to-MSMT. Similarly, we also conduct experiments the codes provided by MMT \cite{ge2020mutual} on MSMT-to-Market and MSMT-to-Duke, respectively. The selected $k$ values of $k$-means are $500$ and $700$, respectively.
For vehicle re-ID, because the experiment results are not reported in MMT \cite{ge2020mutual}, we run its provided codes to get results. The adopted clustering method is DBSCAN \cite{ester1996density}.
For testing, the WaveBlock is not needed.\par

\subsection{Comparison with State-of-the-Arts}
To prove the superiority of the AWB under the MMT \cite{ge2020mutual} framework, we compare the proposed model with state-of-the-art methods on six person re-ID domain adaptations tasks. The comparison results are shown in Table \ref{SOTA}. In terms of mAP, we gain a $9.8\%$, $5.8\%$, $6.2\%$, $6.1\%$, $6.4\%$ and $10.0\%$ improvement on Duke-to-Market, Market-to-Duke, Duke-to-MSMT, Market-to-MSMT, MSMT-to-Duke, and MSMT-to-Market, respectively. As for rank-1, $5.8\%$, $5.8\%$, $10.9\%$, $8.6\%$, $6.3\%$, and $6.7\%$ improvements are obtained, respectively. We attribute the improvement in performance to two aspects. On one hand, the WaveBlocks enhance the complementarity and thus the two networks will not be misled by the same kind of noise to some extent when compared with MMT \cite{ge2020mutual}. On the other hand, the attention modules in AWB learn discriminative and more complementary information which is essential to the performance improvement. In fact, although domain adaptive person re-ID has been explored in many papers, the same experiment setting for vehicle re-ID has not attracted much attention until now. Therefore, we also evaluate it on the vehicle re-ID and image classification tasks. For comparison, we implement the state-of-the-arts algorithm MMT \cite{ge2020mutual} on three vehicle re-ID datasets, i.e. VehicleID \cite{liu_2016deep}, VeRi-776 \cite{liu2016deep}, and VehicleX \cite{naphade20204th}. Two tasks, VehicleID-to-VeRi-776 and VehicleX-to-VeRi-776, are explored. The results are shown in Table \ref{SOTA_ve}. When compared with MMT \cite{ge2020mutual}, as for mAP, $1.0\%$ and $2.0\%$ improvements are achieved, respectively. As for rank-1, we gain a $2.8\%$ and $4.7\%$ improvement, respectively. \par

To prove the generality of the proposed method, we also apply the proposed AWB to image classification task on ImageNet \cite{deng2009imagenet}. The selected baseline is Deep Mutual Learning (DML) \cite{zhang2018deep}. DML \cite{zhang2018deep} designs two neural networks to learn collaboratively and teach each other. For more details, please refer to \cite{zhang2018deep}. The intuition is the same, i.e. using WaveBlock to enhance the complementarity of the two neural networks, and therefore the co-teaching process will be better. We reproduce the experiment results in DML  \cite{zhang2018deep} by using four GPUs. The selected backbones are MobileNet \cite{howard2017mobilenets}. WaveBlocks are arranged after the feature extraction layers. The backbones are trained for $10$ epochs before plugging WaveBlocks. The experimental results are shown in Table \ref{cl}. As for the top-$1$ accuracy, WaveBlock improves the performance without any parameters increasing. If using AWB with I-CBAM or Non-local, the top-$1$ accuracy can be further improved.


\begin{table*}
\caption{The ablation studies about each components in our proposed methods. ``O-CBAM" denotes the original CBAM is used while ``I-CBAM" denotes the improved CBAM is used.  } 
\label{ABLL}
\vspace*{2mm}
\small
  \begin{tabularx}{\hsize}{p{4.5cm}|YYYY|YYYY}
    \hline
    \multicolumn{1}{c|}{\multirow{2}{*}{Methods}} &
    \multicolumn{4}{c|}{Duke-to-Market} &
    \multicolumn{4}{c}{Market-to-Duke} \\
    \cline{2-9}
      & mAP & rank-$1$ & rank-$5$ & rank-$10$ & mAP & rank-$1$ & rank-$5$ & rank-$10$ \\
    \hline\hline
WaveBlock&$76.3$&$90.9$&$96.7$&$97.7$&$68.6$&$82.4$&$91.4$&$94.0$\\

DropBlock&$51.8$&$69.9$&$87.3$&$92.2$&$56.1$&$69.9$&$83.5$&$88.1$\\
\hline
O-CBAM&$76.2$&$90.6$&$97.0$&$98.0$&$67.2$&$80.0$&$90.7$&$93.5$\\

Post-A (O-CBAM)&$75.1$&$89.3$&$96.0$&$97.3$&$64.4$&$77.7$&$89.4$&$92.2$\\

Pre-A (O-CBAM)&$78.7$&$92.5$&$97.1$&$98.2$&$69.0$&$82.4$&$90.9$&$93.5$\\
\hline
I-CBAM&$75.4$&$90.1$&$96.1$&$97.4$&$67.8$&$80.2$&$90.3$&$93.3$\\

Post-A (I-CBAM)&$77.3$&$90.8$&$96.7$&$97.9$&$68.7$&$81.4$&$90.9$&$93.2$\\

Pre-A (I-CBAM)&$78.8$&$92.2$&$97.1$&$98.1$&$70.0$&$82.9$&$91.4$&$93.9$\\
\hline
Non-local&$79.0$&$92.5$&$97.2$&$98.5$&$69.6$&$82.4$&$91.0$&$93.8$\\

Post-A (Non-local)&$\textbf{81.0}$&$\textbf{93.5}$&$\textbf{97.4}$&$\textbf{98.3}$&$\textbf{70.9}$&$\textbf{83.8}$&$\textbf{92.3}$&$\textbf{94.0}$\\

Pre-A (Non-local)&$79.5$&$93.1$&$97.2$&$98.4$&$70.1$&$82.5$&$91.2$&$93.4$\\
    \hline
  \end{tabularx}
  \\
\end{table*}

\subsection{Parameter Analysis and Ablation Studies}
To prove the efficacy of each component in the AWB, we conduct parameter analysis and ablation experiments on DukeMTMC to Market-1501 and Market-1501 to DukeMTMC tasks. The experimental results and analyses are reported below.\par 
\textbf{Selection for the waving width rate $r_w$ and waving height rate $r_h$.}\par 
In the proposed WaveBlock, the waving height rate and waving width rate are of great importance. The mAP and rank-$1$ performance of the different combinations of them are shown in Fig. \ref{para}. The waving height rate is set as $0.5$, $1$, $1.5$, $2.5$, and $3$, while setting the waving width rate as $0.1$, $0.2$, $0.3$, $0.4$, $0.5$, and $0.6$. Apparently, when the waving height rate is $1$, whichever the waving width rate is chose, the experiment setting is same as MMT \cite{ge2020mutual}. For Duke-to-Market, the best performance is achieved when the waving height rate is $1.5$ and waving width rate is $0.3$. Under this setting, the mAP is $76.3\%$ and the rank-$1$ is $90.9\%$. For Market-to-Duke, when the waving height rate equals to $1.5$ and the waving width rate equals to $0.2$, the highest performance, i.e. $68.6\%$ mAP and $82.4\%$ rank-$1$, is achieved. It can be found that the optimal waving height rate is $1.5$ and the optimal waving width rate is $0.2$ to $0.3$. Further, when the waving height rate is larger than $1$, although there is difference in performance, WaveBlocks improve the baseline continuously without any parameters increasing. Then, we discuss why the performance is much better when waving height rate is larger than $1$. According to the gradient-weighted class activation maps, when training a network, the model can focus on informative parts of a feature map. That is why the neural network can accomplish classification, detection, person re-ID, and other tasks. The focusing of a neural network can be understood as a feature map multiplying a mask, where the importance of a pixel is quantified as a number from $0$ to $1$. In the proposed WaveBlock, the waving height rate larger than $1$ means ``highlighting'' while the waving height rate smaller than $1$ means ``ignoring''. Therefore, when the waving height rate is larger than $1$, with the complementarity enhanced, the network can still learn informative parts from highlighting parts because no information is lost and the network can ``choose'' the informative ones from all highlighting parts by multiplying a $0\sim1$ mask. On the contrary, if the weight height is smaller than $1$, though complementarity is enhanced, some parts of features are ignored/deleted. Therefore, though the network has the ability to focus on informative parts, it is impossible to learn from nothing.  The loss of informative features leads to a poor performance. As a result, highlighting certain block is better than ignoring certain block.

\textbf{Effectiveness of the WaveBlock Design.}\par 
To illustrate the effectiveness of the WaveBlock design, the WaveBlock is replaced with the feature dropping block in \cite{dai2019batch}. Also, to avoid disturbance, no attention mechanism is used. The experiment results are reported in Table \ref{ABLL} as ``WaveBlock" and ``DropBlock", respectively. Compared to WaveBlock, for Duke-to-Market, the mAP decreases by $24.5\%$
and the rank-$1$ decreases by $21.0\%$; for Market-to-Duke, the mAP decreases by $12.5\%$ and the rank-$1$ decreases by $12.5\%$. The reason is that DropBlock drops some discriminative and important features, which prevents the two neural networks from fitting training data well. In contrast, the proposed WaveBlock modulates a given feature map with preserved original feature to some degree.

\textbf{Comparison between the original CBAM and improved CBAM.}\par 
The original CBAM (O-CBAM) and improved CBAM (I-CBAM) are compared from two aspects. Firstly, we compare the parameter numbers of the model. For MMT \cite{ge2020mutual}, its backbone ResNet-50 has $23.51$ million parameters. When the backbone is integrated with O-CBAM, it has $26.04$ million parameters. If I-CBAM is used to replace O-CBAM, the parameter numbers decrease to $23.68$ million. I-CBAM only has $0.7\%$ more parameters than backbone while O-CBAM increases the parameters by $10.8\%$. In conclusion, I-CBAM achieves a truly negligible increase in parameters. From the performance aspect, the experimental results are shown in Table \ref{ABLL} as O-CBAM and I-CBAM respectively. When only CBAM is used, I-CBAM achieves competitive mAP and rank-1 performance with O-CBAM both in Duke-to-Market and Market-to-Duke tasks. In the Post-A strategy, we observe performance degradation for O-CBAM in two tasks. Meanwhile, both the Post-A and Pre-A strategies with I-CBAM outperform I-CBAM and the series of O-CBAM. Because WaveBlocks enhance the complementarity of two networks, Post-A and Pre-A perform better. The positions of I-CBAM and WaveBlocks are the same while the ones of O-CBAM and WaveBlocks are different, therefore the former is more effective for focusing on different and discriminative features.

\textbf{Effectiveness of AWB.}\par 
In this part, we try to prove the effectiveness of the attention mechanism in the AWB. Further, two combination strategies for two kinds of attention mechanisms are compared. The experimental results are displayed in Table \ref{ABLL} as I-CBAM and Non-local, respectively. As can be observed, for I-CBAM, Pre-A combination strategy is better than Post-A. It is because the border of the waved feature maps may affect the convolution computing for spatial attention, and the Pre-A strategy avoids this problem. For Non-local block, the performances of both combination strategies are better than adding Non-local block directly. Specifically, the Post-A strategy is much better because directly applying attention modules on waved feature maps is more efficient to produce different and discriminative features and non-local operation reduces the impact of waved regions.





\begin{table*}
\setlength{\abovecaptionskip}{0.3cm}
\setlength{\belowcaptionskip}{0.15cm}
\footnotesize
\centering
\caption{The average differences of two networks in Frobenius norm. ``Baseline" denotes single network,  i.e.  \ the difference is $0$. ``Attention" or ``WaveBlock" denotes only attention mechanism or WaveBlock is used. ``WaveBlock-S" denotes the same $X$ is generated for the two networks. ``AWB" denotes the combination of attention mechanism and WaveBlock.} 
\scalebox{0.9}{
\begin{tabular}{|c|c|c|c|c|c|c|c|c|c|c|c|c|}
\hline
\multirow{2}{*}{Method} &
\multicolumn{6}{c|}{\multirow{1}{*}{Duke-to-Market}} &
\multicolumn{6}{c|}{\multirow{1}{*}{Market-to-Duke}} \\
\cline{2-13}
  &Baseline& MMT \cite{ge2020mutual}&{WaveBlock-S}&WaveBlock&Attention&AWB&Baseline&MMT \cite{ge2020mutual}&WaveBlock-S&WaveBlock&Attention&AWB  \\
\hline
Difference&$0$&$3.14$&$2.41$&$3.31$&$2.94$&$4.74$&$0$&$2.69$&$2.14$&$2.79$&$2.57$&$2.99$\\
\hline
mAP&$53.5$&$71.2$&$72.2$&$76.3$&$79.0$&$81.0$&$48.2$&$65.1$&$66.4$&$68.6$&$69.6$&$70.9$\\
\hline
\end{tabular}
}
\label{ABL-5}
\end{table*}

\begin{table*}
\centering
\caption{The performance of the proposed AWB with different backbones. AWB is a plug-and-play method, which outperforms MMT \cite{ge2020mutual} continuously.} 
  \begin{tabular}{|*{10}{c|}}
  \hline
    \multicolumn{2}{|c|}{\multirow{2}{*}{Method}}  & \multicolumn{4}{c|}{Duke-to-Market} & \multicolumn{4}{c|}{Market-to-Duke}\\\cline{3-10}
    \multicolumn{2}{|c|}{}   & mAP & rank-$1$ & rank-$5$ & rank-$10$ & mAP & rank-$1$ & rank-$5$ & rank-$10$ \\\hline
    \multirow{4}{*}{WideResNet-50 \cite{zagoruyko2016wide}} & MMT \cite{ge2020mutual} & $57.5$ & $78.5$ & $91.5$ & $94.5$ & $60.1$ & $74.8 $ & $ 86.8$ & $ 90.4$ \\\cline{2-2}
        & WaveBlock & $63.8$ &$84.8$ & $94.1$ & $96.2$ & $62.9$ & $77.6$ & $87.7$ & $91.1$ \\\cline{2-2}
        & Pre-A (I-CBAM) & $66.5$ &$86.0$ & $94.7$ & $96.7$ & $66.4$ & $80.9$ & $89.6$ & $92.1$ \\\cline{2-2}
        & Post-A (Non-local) & $\textbf{73.7}$ & $\textbf{89.8}$ & $\textbf{95.9}$ & $\textbf{97.3}$ & $\textbf{67.2}$ & $\textbf{79.9}$ & $\textbf{90.2}$ & $\textbf{92.8}$ \\\hline
    \multirow{4}{*}{DenseNet-121 \cite{huang2017densely}} & MMT \cite{ge2020mutual} & $73.7$ & $89.2$ & $96.1$ & $97.4$ & $65.5$ & $78.1$ & $ 89.5$ & $92.9$ \\\cline{2-2}
        & WaveBlock & $77.7$ & $91.1$ & $96.7$ & $97.8$ & $69.7$ & $ 81.3$ & $90.8$ & $ 93.1$ \\\cline{2-2}
        & Pre-A (I-CBAM) & $78.4$ & $91.4$ & $96.8$ & $97.8$ & $70.1$ & $82.0$ & $91.1$ & $92.9$ \\\cline{2-2}
        & Post-A (Non-local) & $\textbf{82.9}$ & $\textbf{92.5}$ & $\textbf{97.5}$ & $\textbf{98.6}$ & $\textbf{72.5}$ & $\textbf{83.6}$ & $\textbf{91.9}$ & $\textbf{93.9}$ \\\hline
  \end{tabular}
\label{backbone}
\end{table*}

\subsection{Further Discussion}
\textbf{The relationship between the created difference and the performance}\par 
First, the difference created by WaveBlocks and enlarged by attention mechanism is quantified in this part.  Further, the relationship between the quantified difference and the performance is discussed. The difference is quantified by calculating the Frobenius norm between two gradient-weighted class activation maps \cite{selvaraju2017grad} of the same input after Stage $3$ or the proposed modules, as illustrated in the introduction section. Further, the differences in the Frobenius norm for all images are averaged to obtain final quantified differences.  The quantified difference and the corresponding mAP are shown in Table  \ref{ABL-5}.  
We have the following settings: Baseline (single network), MMT, WaveBlock with the same X, WaveBlock, Attention, and AWB.  `Baseline' only uses a single network, therefore, the difference can be regarded as $0$.  `MMT' uses two networks, therefore, the difference is created to some degree, and complementarity is enhanced to some extent. Comparing with the baseline, performance gains significant improvement.  `WaveBlock with the same X' denotes the same shape of WaveBlocks is adopted for both networks, i.e. when generating the waving random integer, $X_1$ is always equal to $X_2$.  `Attention' denotes the strategy of Post-A with Non-local. For Duke-to-Market and Market-to-Duke, the mAPs are $72.2\%$ and $66.4\%$, respectively. However, for MMT, the two networks are ``independent", while the same X gives them a ``bind", therefore the created difference decreases. For WaveBlock, the difference between the two networks is enlarged,  and the performance gains significant improvement. Although only using the attention mechanism is not beneficial for creating a difference, the model can learn more discriminative feature, therefore performance improvement can still be observed.  Finally, for the AWB, the attention mechanism enlarges the created difference and discover more discriminative feature, thus, the performance is the best.



\textbf{Stable performance improvement with different backbones.}\par 
To prove that the proposed method is a plug-and-play method, we try some other backbones besides ResNet-50 \cite{he2016deep}. The selected backbones are WideResNet-50 \cite{zagoruyko2016wide} and DenseNet-121 \cite{huang2017densely}. Similar with the modification for ResNet-50 \cite{he2016deep}, i.e. the last spatial down-sampling operation is removed, we also modify  WideResNet-50 \cite{zagoruyko2016wide} and DenseNet-121 \cite{huang2017densely} to obtain a higher resolution. Specifically, the modification for WideResNet-50 \cite{zagoruyko2016wide} is same as ResNet-50 \cite{he2016deep} and the average pooling operation in the last transition layer of DenseNet-121 \cite{huang2017densely} is removed. \par 
For WideResNet-50 \cite{zagoruyko2016wide}, we plug the proposed WaveBlocks or AWBs after the stage $2$ and $3$. For DenseNet-121 \cite{huang2017densely}, they are arranged after the Dense Block $2$ and $3$. The experiment results are shown in Table \ref{backbone}. In Duke-to-Market task, the mAP increases by $16.2\%$ with WideResNet-50 \cite{zagoruyko2016wide} and increases by $9.2\%$ with DenseNet-121 \cite{huang2017densely}. Also, in Market-to-Duke task, the mAP increases by $7.1\%$ and $7.0\%$ with two backbones, respectively. Further, we achieve higher mAP and rank-$1$ performance in two tasks by using DenseNet-121 \cite{huang2017densely} than ResNet-50 \cite{huang2017densely} as our backbone.

\begin{table*}
\centering
\caption{The performance of the proposed AWB under different $k$ values. AWB outperforms MMT \cite{ge2020mutual} continuously.} 
  \begin{tabular}{|*{10}{c|}}\hline
    \multicolumn{2}{|c|}{\multirow{2}{*}{Method}}  & \multicolumn{4}{c|}{Duke-to-Market} & \multicolumn{4}{c|}{Market-to-Duke}\\\cline{3-10}
    \multicolumn{2}{|c|}{}   & mAP & rank-$1$ & rank-$5$ & rank-$10$ & mAP & rank-$1$ & rank-$5$ & rank-$10$ \\\hline
    \multirow{3}{*}{500} & MMT \cite{ge2020mutual} & $71.2$ & $87.7$ & $94.9$ & $96.9$ & $63.1$ & $76.8$ & $88.0$ & $92.2$ \\\cline{2-2}
        & Pre-A (I-CBAM) & $78.8$ &$92.2$ & $97.1$ & $98.1$ & $66.3$ & $80.2$ & $89.9$ & $92.3$ \\\cline{2-2}
        & Post-A (Non-local) & $\textbf{81.0}$ & $\textbf{93.5}$ & $\textbf{97.4}$ & $\textbf{98.3}$ & $67.5$ & $80.9$ & $90.6$ & $93.0$ \\\hline
    \multirow{3}{*}{700} & MMT \cite{ge2020mutual} & $69.0$ & $86.8$ & $94.6$ & $96.9$ & $65.1$ & $78.0$ & $88.8$ & $92.5$ \\\cline{2-2}
        & Pre-A (I-CBAM) & $76.8$ & $91.5$ & $97.1$ & $98.2$ & $70.0$ & $82.9$ & $91.4$ & $93.9$ \\\cline{2-2}
        & Post-A (Non-local) & $78.8$ & $93.0$ & $97.2$ & $98.3$ & $\textbf{70.9}$ & $\textbf{83.8}$ & $\textbf{92.3}$ & $\textbf{94.0}$ \\\hline
    \multirow{3}{*}{900} & MMT \cite{ge2020mutual} & $66.2$ & $86.8$ & $94.9$ & $96.6$ & $63.1$ & $77.4$ & $88.1$ & $92.5$ \\\cline{2-2}
        & Pre-A (I-CBAM)& $75.0$ & $91.9$ & $97.1$ & $98.1$ &$68.2$ & $81.6$ & $91.2$ & $93.7$ \\\cline{2-2}
        & Post-A (Non-local) & $73.8$ & $91.4$ & $96.5$ & $97.7$ & $69.5$ & $81.8$ & $91.3$& $93.5$ \\\hline
  \end{tabular}
\label{diffk}
\end{table*}

\textbf{Stable performance improvement with different $k$ values.}\par 
Actually, the AWB can improve performance with different $k$ values stably. Similar with MMT \cite{ge2020mutual}, we have tried three different $k$, i.e. $500$, $700$, and $900$, for Duke-to-Market and Market-to-Duke tasks, respectively. The experiment results are shown in Table \ref{diffk}. No matter which strategy (Pre-A (I-CBAM) or Post-A (Non-local)) is chosen, the performance is improved significantly when compared with MMT \cite{ge2020mutual}. These experiment results prove the generality of the proposed method. \par

\begin{table*}
\caption{The experimental results of state-of-the-art self-supervised learning methods on UDA person re-ID tasks. It can be found that them cannot handle the re-ID tasks well. `-' denotes a non-convergence result is observed. The implementation is based on the authors’ code.} 
\label{ssl}
\vspace*{2mm}
\small
  \begin{tabularx}{\hsize}{p{4cm}|YYYY|YYYY}
    \hline
    \multicolumn{1}{c|}{\multirow{2}{*}{Methods}} &
    \multicolumn{4}{c|}{Duke-to-Market} &
    \multicolumn{4}{c}{Market-to-Duke} \\
    \cline{2-9}
      & mAP & rank-$1$ & rank-$5$ & rank-$10$ & mAP & rank-$1$ & rank-$5$ & rank-$10$ \\
    \hline\hline
SimSiam \cite{chen2021exploring}&$7.7$&$25.3$&$41.9$&$49.8$&$4.8$&$13.5$&$23.8$&$29.9$\\
BYOL \cite{NEURIPS2020_f3ada80d}&$6.6$&$19.2$&$36.1$&$44.7$&$5.5$&$13.1$&$23.7$&$29.9$\\
SwAV \cite{caron2020unsupervised}&$-$&$-$&$-$&$-$&$-$&$-$&$-$&$-$\\
MoCo V2 \cite{he2020momentum}&$8.9$&$24.0$&$38.6$&$46.2$&$5.2$&$9.5$&$18.8$&$24.4$\\
    \hline
WaveBlock&$\textbf{76.3}$&$\textbf{90.9}$&$\textbf{96.7}$&$\textbf{97.7}$&$\textbf{68.6}$&$\textbf{82.4}$&$\textbf{91.4}$&$\textbf{94.0}$\\
    \hline
  \end{tabularx}
  \\
\end{table*}

\textbf{Comparison with self-supervised learning methods.}\par 
The state-of-the-art self-supervised learning methods are implemented to person re-ID tasks in this part. Specifically, we choose SimSiam \cite{chen2021exploring}, BYOL \cite{NEURIPS2020_f3ada80d}, SwAV \cite{caron2020unsupervised}, and MoCo V2 \cite{he2020momentum}.  To mimic UDA, the experiment setting is using the pre-trained model on the source domain to conduct self-supervised learning on the target domain. However, although we try our best to adjust the hyper-parameters carefully, all of the self-supervised learning methods cannot work well on person re-ID tasks. This phenomenon is also observed in SpCL \cite{ge2020selfpaced} and \cite{DBLP:journals/corr/abs-2010-07608} independently.  The experimental results are shown in Table \ref{ssl}. We point out the reasons as follows. First, the scale of commonly used person re-ID datasets is not applicable for the above self-supervised methods. The most common dataset for the above self-supervised methods is ImageNet, which has $1.2$ million images for training. However, the common person re-ID datasets, such as DukeMTMC \cite{zheng2017unlabeled} and Market1501 \cite{zheng2015scalable}, only have about $10,000 \sim 20,000$ images. The scale of ImageNet is nearly $100$ times larger than the person re-ID datasets. As far as we concern, to conduct the above self-supervised learning, the scale of training datasets is crucial to learn discriminative features.  Then, the discriminative features needed by the person re-ID tasks cannot be learned well by the above self-supervised methods directly. In this part, we use MoCo V2 for example to analyze and conduct experiments. MoCo V2 tries to learn ``instance discrimination'', i.e., using one positive to conduct contrastive learning and the learned features are ``sparse'' in the feature space. The learned instance discrimination is suitable for downstream tasks. For directly applying the sparse feature to person re-ID tasks, when using Euclidean distance to test, the performance of sparse features is unsatisfactory because the core of re-ID tasks is to encode and model intra-/inter-class variations \cite{ge2020selfpaced}. However, by carefully designing the hyper-parameters, pre-training a model using MoCo V2  \cite{he2020momentum} is beneficial for person re-ID tasks. Specifically, we collect a large-scale private person re-ID dataset, which has about $600$K IDs and $28$ million images. Then, we adjust the hyper-parameters and pre-train the MoCo V2  \cite{he2020momentum}  on that large-scale dataset (without using ImageNet). Using the unsupervised pre-trained model to initialize a network, we observe the triple loss and the identification loss going down quickly. Also, the network reaches very high performance in a short time. The final performance (mAP) is better than using ImageNet to initialize the model. In conclusion, the self-supervised method is useful for person re-ID tasks to some degree.



\section{Conclusion}
In this paper, a parameter-free module, the WaveBlock, is first proposed. Then, we design two kinds of combination strategies, \textit{i.e.} pre-attention and post-attention, to integrate the proposed WaveBlock with the attention mechanism. We use the WaveBlock to create a difference between features learned by two networks under the framework of MMT. An attention mechanism is also utilized to enlarge the difference and learn different and discriminative features on the basis of WaveBlock. By plugging the proposed AWB into the MMT, the complementarity of the two networks is enhanced and the possibility of their being biased towards the same kind of noise is decreased. Extensive experiments show that the proposed AWB under the MMT framework outperforms the state-of-the-art unsupervised domain adaptation person re-identification methods by a large margin. Further, the generality of the proposed method is proved by applying it a vehicle re-identification and image classification tasks.

\section*{Acknowledgment}
We thank Informatization Office of Beihang University for the supply of High Performance Computing Platform. We also would like to thank Anna Hennig who helped proofreading the paper. Wenhao Wang wants to thank Jin Fan and Bo Qin for their generous computer technique support.

\bibliographystyle{plain}
\bibliography{mylib}
\end{document}